\documentclass[a4paper,UKenglish]{lipics-v2018}

\usepackage[justification=centering]{caption}

\usepackage{microtype}

\title{Analysis of Irregular Spatial Data with Machine Learning: Classification of Building Patterns with a Graph Convolutional Neural Network}

\titlerunning{Analysis of Irregular Spatial Data with Machine Learning}

\author{Xiongfeng Yan}{School of Resource and Environmental Sciences, Wuhan University, 430079 Wuhan, China}{xiongfeng.yan@whu.edu.cn}{https://orcid.org/0000-0003-4748-464X}{}

\author{Tinghua Ai}{School of Resource and Environmental Sciences, Wuhan University, 430079 Wuhan, China}{tinghuaai@whu.edu.cn}{https://orcid.org/0000-0002-6581-9872}{}

\authorrunning{X.\, Yan and T.\, Ai}

\Copyright{}

\subjclass{Information systems $\rightarrow$ Geographic information systems}


\category{arXiv preprint}

\relatedversion{This work was published in \emph{Proceedings of 10th International Conference on Geographic Information Science}, which is available at http://www.dagstuhl.de/dagpub/978-3-95977-083-5. In this version, we corrected several mathematical symbols in the formulas.}

\supplement{}

\funding{This work was supported by the National Natural Science Foundation of China [41531180] and the International Exchange Program for Graduate Students of Wuhan University.}

\acknowledgements{}

\EventEditors{Stephan Winter, Amy Griffin, and Monika Sester}
\EventNoEds{3}
\EventLongTitle{10th International Conference on Geographic Information Science (GIScience 2018)}
\EventShortTitle{GIScience 2018}
\EventAcronym{GIScience}
\EventYear{2018}
\EventDate{August 28--31, 2018}
\EventLocation{Melbourne, Australia}
\EventLogo{}
\SeriesVolume{114}
\ArticleNo{96}
\nolinenumbers 
\hideLIPIcs  

\begin{document}

\maketitle

\begin{abstract}
Machine learning methods such as convolutional neural networks (CNNs) are becoming an integral part of scientific research in many disciplines, spatial vector data often fail to be analyzed using these powerful learning methods because of its irregularities. With the aid of graph Fourier transform and convolution theorem, it is possible to convert the convolution as a point-wise product in Fourier domain and construct a learning architecture of CNN on graph for the analysis task of irregular spatial data. In this study, we used the classification task of building patterns as a case study to test this method, and experiments showed that this method has achieved outstanding results in identifying regular and irregular patterns, and has significantly improved in comparing with other methods.
\end{abstract}

\section{Introduction}

With the improvement of computing power and the advent of the data era, machine learning methods are becoming an integral part of scientific research in many disciplines. As a supervised learning method, CNN has excellent performance in many fields, such as computer vision and natural language processing. These successes are mainly attributed to its two important properties: first, inspired by neuronal processing, the CNN focuses on local structures (Local Receptive Fields, LRFs) and combines them into a whole, which can be applied to classification or recommendation tasks. Second, local structures of different regions can be detected by using the same convolution kernels, that is, weight sharing. The former is accordance with the compositionality of objects and the hierarchy of cognition, and the latter reduces the complexity and improves the learnability.

However, it should be noted that both the local connection and weight sharing require that the regularity of input data; in other word, the LRFs are fixed, normalized, and can be clearly defined. For example, the image in visual analysis is processed by pixels that are organized into a grid, and the sentence in natural language processing is processed by words that are organized into a linear arrangement. However, for most of the spatial vector data in GIS fields, the arrangements, combinations, or connections between objects could be more diversified, and it is often difficult to satisfy this requirements of specification. Therefore, this kinds of data cannot directly use these powerful learning methods.

Although most spatial vector data cannot be organized according to a structure that satisfies the regularity (such as a grid or array), it is still possible to modeled by a graph structure. The graph cannot define a convolution operation in the vertex domain directly, but in virtue of graph Fourier transform and convolution theorem, the operation can be transformed into a point-wise product in the Fourier domain, which is similar to the transformation of spatial domain convolution into frequency domain convolution in image processing. Based on this idea, we propose a learning architecture of CNN on graph, which we term GCNN, for the analysis tasks of irregular spatial data.

In this study, we focus on using the GCNN to solve the classification problem of building group patterns, which plays an important role in various applications, such as urban morphology and map generalization. Although the related researches have been carried out for decades, there are still some problems such as incomplete taxonomy and inconsistent recognition rules. The introduction of machine learning method could be an effective attempt and could supplement attempts to solve such classical problems in spatial analysis. In the following sections, we will describe detailed methods, then conduct experiments and compare with other similar methods and, finally, discuss and conclude this study.

\section{Methodology}

\subsection{Definition of Building Pattern Classification}

Building patterns refer to visually salient structures exhibited collectively by a group of buildings\cite{Du2015Representation}. Traditional patterns detection methods are to predefine some specific perceptual rules according to the characteristics of azimuth angle, direction difference and proximity, and then to inquire whether there is a local group that satisfy such rules\cite{Du2016Extracting}\cite{Wei2018On}\cite{Zhang2013Building}. But these rules are difficult to formalize and too rigid, which inevitably lead to an unsatisfactory result\cite{He2018Recognition}.

Similar to image processing, determining which pattern a building group visually belongs to is essentially an issue of classification. A building group is an analogy to an image, and each building is analogous to a pixel, and its semantic attributes and shape features are analogous to the color channels.

\subsubsection{Features of individual buildings}

Individual building has spatial features that describe its graphical structures and semantic features that describe its attributes, which in combination can effectively reflect its basic form. For the description of these features, dozens of indices have been proposed\cite{Wei2018On}. In this study, we mainly consider five indices of the area $A_b$, main direction $\alpha $, and three shape indices including length-width ratio $R_{lw}$, area ratio $R_A$, and compactness $C$ as input features for individual buildings. These indices are illustrated in Fig. \ref{figure 1}.

\subsubsection{Graph representation of building group}

Graph is an ideal tool to describe the relationships between multiple objects. Delaunay triangulation (DT) and Minimum Spanning Tree (MST) are the two most commonly used ways due to they can take spatial constraints and other contextual constraints into consideration, such as proximity.

\begin{figure}
  \centering
  \begin{minipage}[h]{2.62in}
    \centering
    \includegraphics[width=2.62in]{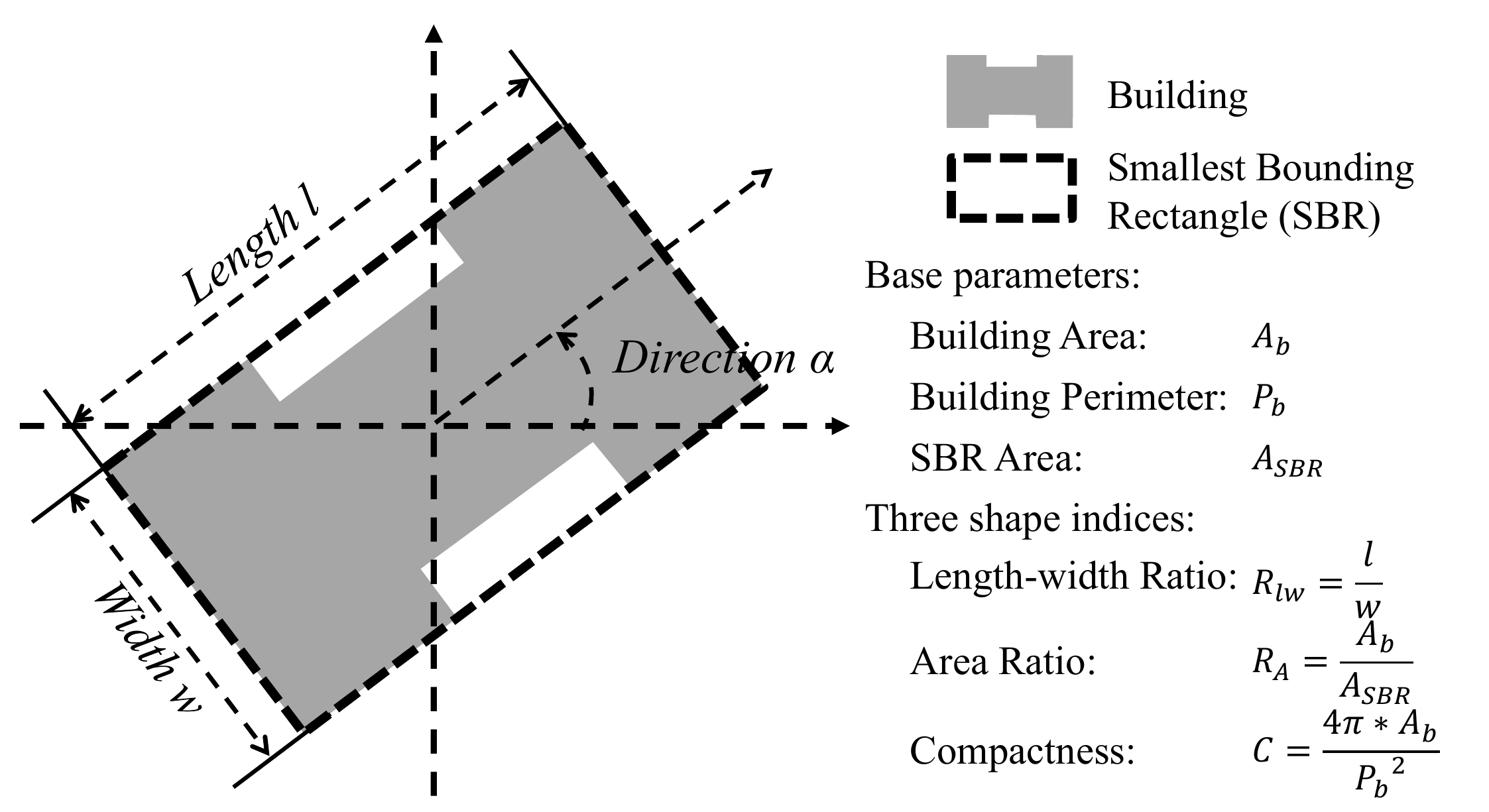}
    \caption{Input feature indices}\label{figure 1}
  \end{minipage}
  \begin{minipage}[h]{2.62in}
    \centering
    \includegraphics[width=2.62in]{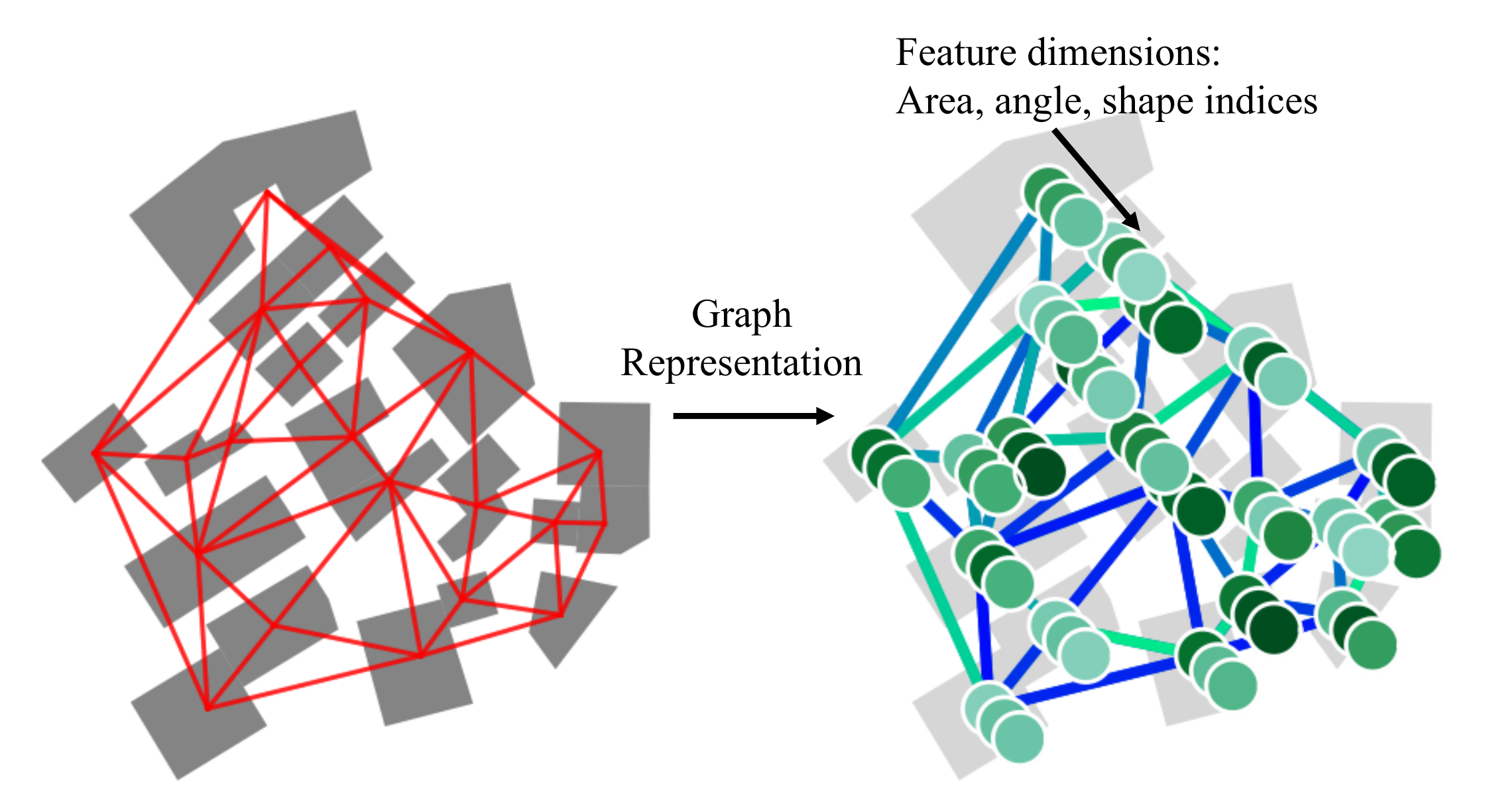}
    \caption{Graph construction}\label{figure 2}
  \end{minipage}
\end{figure}

Regardless of whether DT or MST, they can be defined as $\mathcal{G}\mathrm{=(}\mathcal{V},\mathcal{E},\mathcal{W}\mathrm{)}$, where $\mathcal{V}$ and $\mathcal{E}$ is a finite set of $\left|\mathcal{V}\right|=n$ vertices and edges, respectively, $\mathcal{W}\in {\mathbb{R}}^{n\times n}$ is a adjacency matrix encoding the weight between two vertices, and each vertex also contains one or several input features, as seen in Fig. \ref{figure 2}.

\subsection{Graph Convolutional Neural Network}

\subsubsection{Graph Fourier transform}

The Fourier transform is an effective tool in signal analysis and image processing, it decomposes an original function (e.g., a signal or an image) into the frequencies that make it up. The process is essentially a linear transformation by using given orthogonal basics $\left\langle f,e^{i\omega t}\right\rangle $.

For graph-structured data, we utilize the eigenvectors ${\chi }_{\ell }$ of Laplacian as the decomposition basics instead of complex exponentials, then define the graph Fourier transform as:
\begin{equation}\label{equation 1}
  \hat{f}\left({\lambda }_{\ell }\right) = \sum_{n=1}^{N}{{\chi }^T_{\ell }(n)f(n)} = {\mathcal{X}}^Tf
\end{equation}

Where, ${\lambda }_{\ell }$ denotes the eigenvalues and $\mathcal{X}$ denotes their matrix. The inverse Fourier transform is given as:
\begin{equation}\label{equation 2}
  f\left(n\right) = \sum_{\ell=1}^{N}{\hat{f}\left({\lambda }_{\ell }\right){\chi }_{\ell }(n)} = {\mathcal{X}}\hat{f}
\end{equation}

This definition is precise analogous to the classical case, and it can be interpreted as an expansion of $f$ in terms of the eigenvectors of the Laplacian\cite{Hammond2009Wavelets}\cite{Shuman2016Vertex}.

\subsubsection{Convolution on graph}

Because we cannot conduct the convolution in the vertex domain directly, we can attempt to convert this operation into a point-wise product in the Fourier domain by means of graph Fourier transform and convolution theorem, and it can be defined as:
\begin{equation}\label{equation 3}
  f*g = \sum_{\ell=1}^{N}{\hat{f}\left({\lambda }_{\ell}\right){\hat{g}\left({\lambda }_{\ell }\right)\chi }_{\ell }(n)} = \mathcal{X}\left(\left({\mathcal{X}}^Tf\right)\cdot \left({\mathcal{X}}^Tg\right)\right)
\end{equation}

Denoting the transform ${\mathcal{X}}^Tg$ as a set of free parameters $\left\{ \hat{g}\left({\lambda }_1\right),\ \dots ,\hat{g}\left({\lambda }_N\right) \right\}$ in the Fourier domain (i.e. the Eigenspaces of Lalpace), which also can be understood as a function $\hat{g}\left(\mathrm{\Lambda }\right)$ of the eigenvalues, and then, the convolution can be written as:
\begin{equation}\label{equation 4}
  f*g = \mathcal{X}diag\left(\hat{g}\left({\lambda }_1\right),\ \dots ,\hat{g}\left({\lambda }_N\right)\right){\mathcal{X}}^Tf\mathrm{=}\mathcal{X}{\hat{g}}_{\theta }\left(\mathrm{\Lambda }\right){\mathcal{X}}^Tf
\end{equation}

\begin{figure}
  \centering
  \includegraphics[width=4.6in]{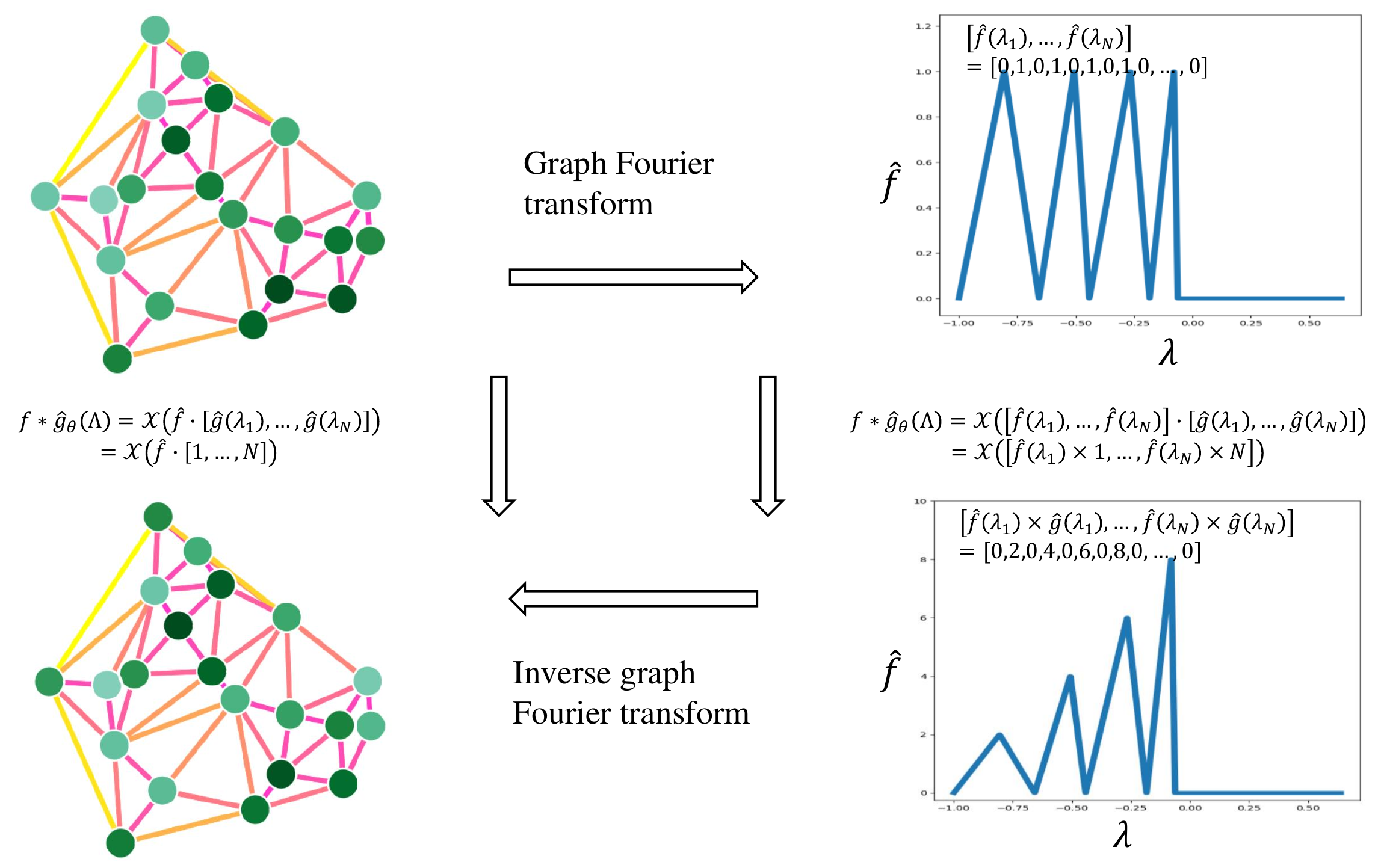}
  \caption{The convolution of a graph $f$ with a kernel of free parameters $\left[1,...,N\right]$}\label{figure 3}
\end{figure}

An illustration of the operation is shown in Fig. \ref{figure 3}.

\subsubsection{Polynomial approximation for fast localized convolution}

The above definition of convolution operation on graph still has two limitations: 1) in each operation, the Eigen decomposition must to be performed, which will have a large amount of computational cost; 2) without considering the locality in space, the feature values of a vertex can be related to the global vertices after this operation, which is not consistent with the local connection property of the classical CNN\cite{Bruna2013Spectral}\cite{Defferrard2016Convolutional}.

In response to these problems, Hammond\cite{Hammond2009Wavelets} proposed a fast localized convolution based on low-order polynomial approximation that represent ${\hat{g}}_{\theta }\left(\mathrm{\Lambda }\right)$ as a polynomial function of the eigenvalues:
\begin{equation}\label{equation 5}
  {\hat{g}}_{\theta }\left(\mathrm{\Lambda }\right) = \sum_{k=0}^{K-1}{{\theta }_k{\mathrm{\Lambda }}^k}
\end{equation}

Then, the Formula \eqref{equation 4} can be rewritten as:
\begin{equation}\label{equation 6}
  f*g = \mathcal{X}\left(\sum_{k=0}^{K-1}{{\theta }_k{\mathrm{\Lambda }}^k}\right){\mathcal{X}}^Tf = \left(\sum_{k=0}^{K-1}{{\theta }_k\left(\mathcal{X}{\mathrm{\Lambda }}^k{\mathcal{X}}^T\right)}\right)f = \left(\sum_{k=0}^{K-1}{{\theta }_k{\mathcal{L}}^k}\right)f
\end{equation}

As can be seen, there is no need to perform the Eigen decomposition anymore, and the feature values of vertex are related only to its K-order neighboring vertices, which satisfies the locality in space.

\subsubsection{Architecture of convolutional neural network on graph}

\begin{figure}
  \centering
  \includegraphics[width=5.2in]{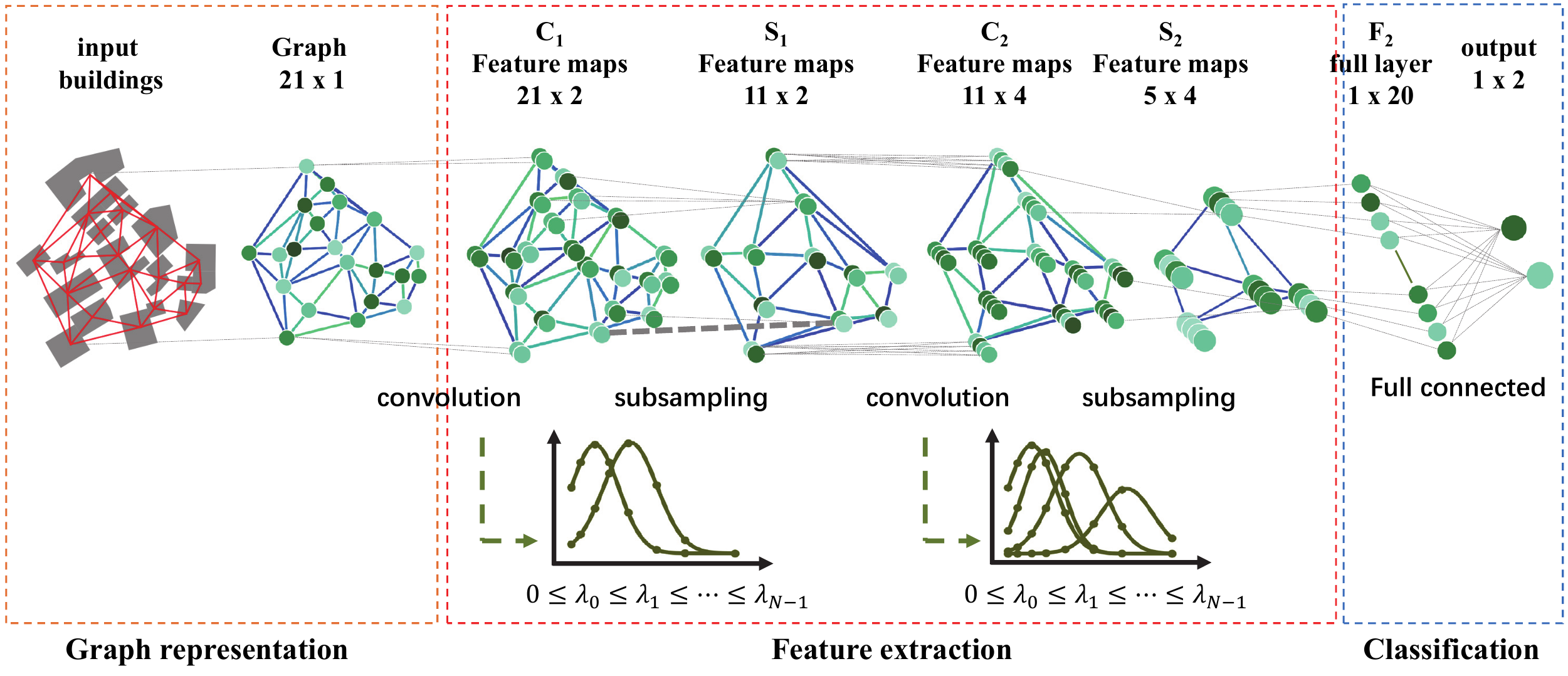}
  \caption{The architecture of convolutional neural network on graph}\label{figure 4}
\end{figure}

Based on the above-defined graph convolution, we propose a learning architecture of CNN on graph for the classification of building patterns, as seen in Fig. \ref{figure 4}. This architecture includes convolutional, subsampling, and full connected layers, where subsampling layer is optional and the full connected layer is the same as the classical CNN. We input a building group that has already been modeled as graph to this architecture, after the steps of feature extraction and classification, we can get the probability that it belongs to each class and choose the class with maximum probability as the final classification result.

\section{Experiments}
The experimental buildings were extracted from a large-scale 1:2000 topological map of the city of Guangzhou, China. We divided them into separated groups by using road network division and simple clustering techniques, and each group contains 20-128 buildings. Then, we manually identified the two patterns, regular and irregular, from all of the groups. Each group was estimated by at least three participants to ensure the correctness, and the ambiguous groups were discarded. Last, there are 2647 and 2646 available groups for regular and irregular pattern, respectively, and they contain a total of 318 598 buildings. Each group can serve as a sample for the GCNN, all samples were split into training, validation and test sets by 6:2:2, and input features of all data were standardized using training set.

We used a shallow GCNN architecture with four convolutional layers and one full connected layer to test the datasets, each convolutional layer contains 24 third-order polynomial convolution kernels. The more convolutional layers, the more complex the model is and the more samples are required. In addition, regularization and dropout techniques are also used to control the complexity, and their parameters are referenced from empirical values and fine-tuned. The accuracy is 98.04\%, which is better than that of SVM\cite{liqiang2013spatial} and random forest\cite{He2018Recognition} methods, the comparison results are shown in Table \ref{table 1}.
\begin{table}
  \centering
  \begin{tabular}
            {|p{0.8in}<{\centering}|p{1.0in}<{\centering}|p{1.0in}<{\centering}|p{1.0in}<{\centering}|} \hline
            Method & SVM & Random Forest & GCNN \\ \hline
            Accuracy & 90.2\% & 93.4\% & 98.04\% \\ \hline
            \end{tabular}
  \caption{Accuracies of the proposed method and other methods}\label{table 1}
\end{table}

\begin{figure}
  \centering
  \includegraphics[width=5.0in]{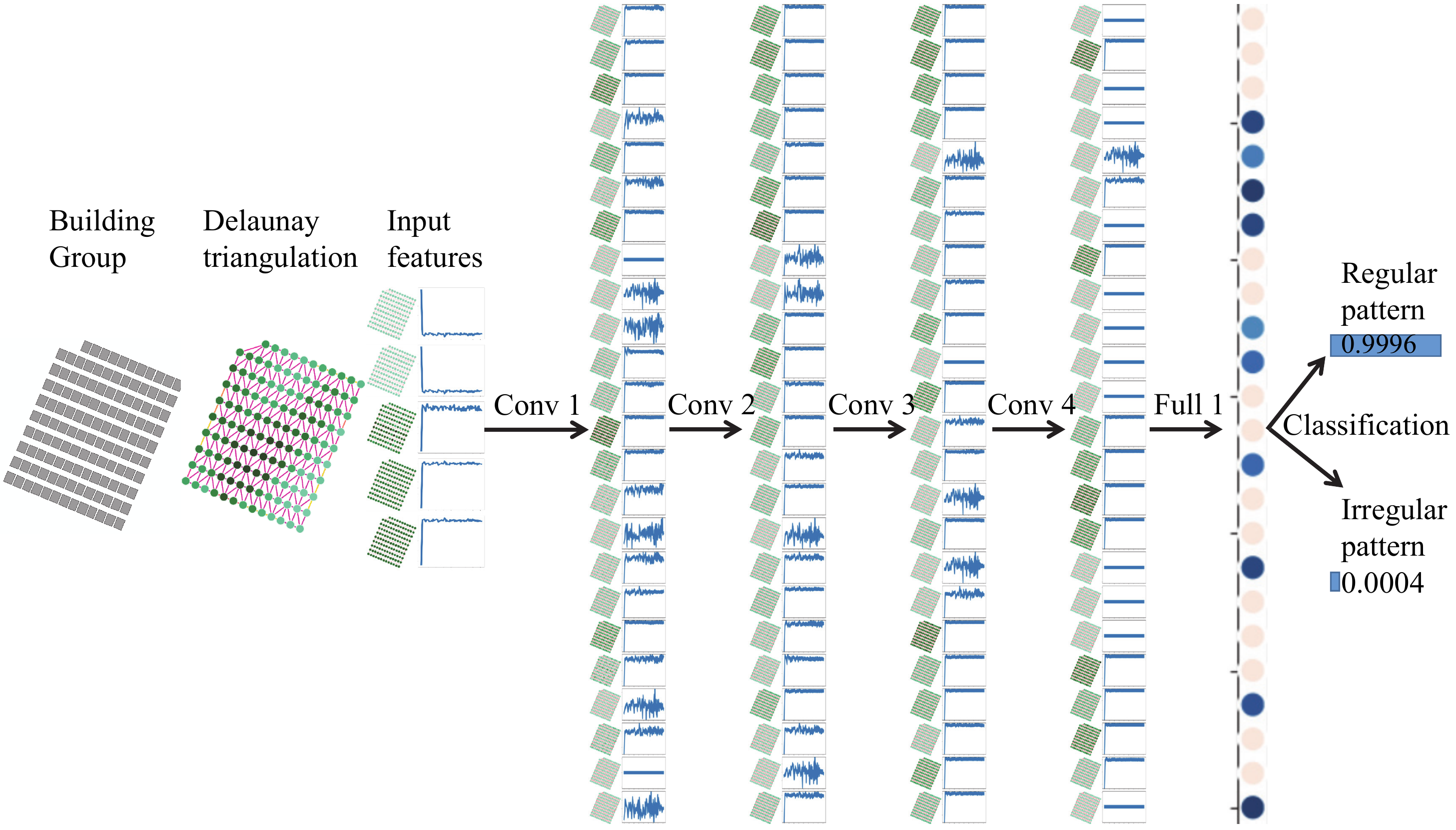}
  \caption{The activations of an example GCNN architecture}\label{figure 5}
\end{figure}

The activation of a sample is shown in Fig. \ref{figure 5} and the input volume stores the graph of building group (left) and the last volume holds the scores for each class (right).

\begin{figure}
  \centering
  \includegraphics[width=5.0in]{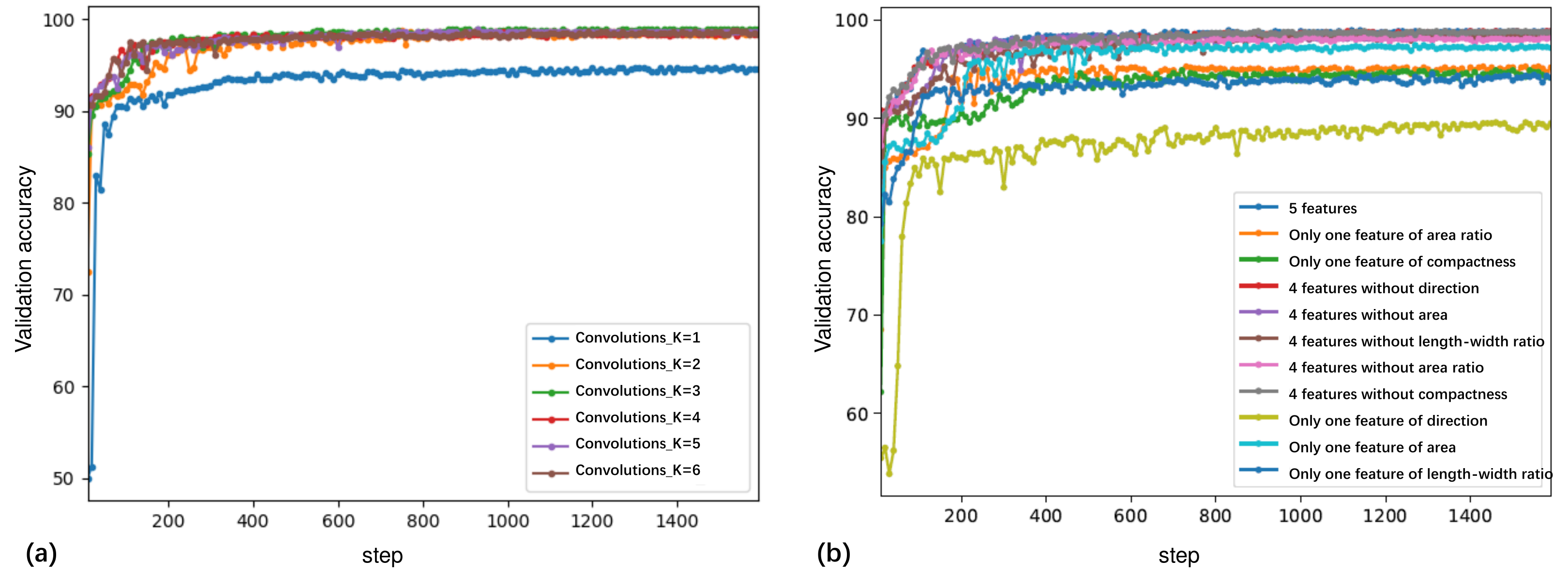}
  \caption{Performances when taking different K values or inputting different features}\label{figure 6}
\end{figure}

In this model, the order K of the polynomial is one of the important parameters. We tested the values, from one to six, and these performances on the validation set are shown in Fig. \ref{figure 6}a. The comparison found that it achieved the best performance when K=3. The larger of K, the more complex of the training and the longer it takes. We further tested the effect of input features of individual building on the classification of group patterns. We tried to train and learn by using only one index at a time or all other indices except for one as input features, these results are shown in Fig. \ref{figure 6}b and we found that the area was an important feature for describing individual buildings and the accuracy reached 96.34\% when only the area index was used. This may be due to the fact that areas of buildings in a regular pattern are more homogeneous.

\section{Discussion and Conclusion}

As a classical problem in the analysis of irregular spatial data, the traditional building pattern classification method needs to manually extract features and design rules for specific patterns. In this paper, we propose a GCNN model in which represent the building group as graph and convert the convolution from the vertex domain into a point-wise product in the Fourier domain. This model can directly extract patterns characteristics based on the training and learning of example data. Experiments showed that proposed method has achieved outstanding results in identifying regular and irregular patterns, and has significantly improved in comparing with other methods. Meanwhile, it has great potential to extend to other analysis tasks of irregular spatial data, such as classification of road patterns and identification of point clouds.

The difficulties of this method lie in the selection of input features and the training process. We have selected five features in our experiments, but there are still many other descriptive indices. Determining which indices can better describe building patterns and how to apply them to the learning model still requires more experiments, and the principal component analysis may be a worthwhile approach to try. The training of GCNN requires a large amount of high-quality examples, otherwise it will easily lead to overfitting, especially for deep networks with many convolutional layers. In the follow-up work, Volunteer Geographic Information (VGI) is a desirable and feasible data source.



\bibliography{lipics-v2018-short-paper-96}

\end{document}